\documentclass[runningheads]{llncs}
\usepackage{biblatex}  
\usepackage{booktabs}
\usepackage{algpseudocode}
\usepackage{algorithm}
\usepackage[T1]{fontenc}
\usepackage{graphicx}
\usepackage{subcaption}
\usepackage{multirow} 
\begin{document}
\title{Pseudo-label Based Domain Adaptation for Zero-Shot Text Steganalysis}
%
\author{Yufei Luo\inst{1}\and
Zhen Yang*\inst{1} \and
Ru Zhang*\inst{1} \and
Jianyi Liu\inst{1}}
\authorrunning{Yufei Luo et al.}
%

\institute{
  Beijing University of Posts and Telecommunications, Beijing, China \\
  \email{yangzhenyz@bupt.edu.cn, zhangru@bupt.edu.cn}
}

\maketitle              
\begin{abstract}
Currently, most methods for text steganalysis are based on deep neural networks (DNNs). However, in real-life scenarios, obtaining a sufficient amount of labeled stego-text for correctly training networks using a large number of parameters is often challenging and costly. Additionally, due to a phenomenon known as dataset bias or domain shift, recognition models trained on a large dataset exhibit poor generalization performance on novel datasets and tasks. Therefore, to address the issues of missing labeled data and inadequate model generalization in text steganalysis, this paper proposes a cross-domain stego-text analysis method (PDTS) based on pseudo-labeling and domain adaptation (unsupervised learning). Specifically, we propose a model architecture combining pretrained BERT with a single-layer Bi-LSTM to learn and extract generic features across tasks and generate task-specific representations. Considering the differential contributions of different features to steganalysis, we further design a feature filtering mechanism to achieve selective feature propagation, thereby enhancing classification performance. We train the model using labeled source domain data and adapt it to target domain data distribution using pseudo-labels for unlabeled target domain data through self-training. In the label estimation step, instead of using a static sampling strategy, we propose a progressive sampling strategy to gradually increase the number of selected pseudo-label candidates. Experimental results demonstrate that our method performs well in zero-shot text steganalysis tasks, achieving high detection accuracy even in the absence of labeled data in the target domain, and outperforms current zero-shot text steganalysis methods.

\end{abstract}
\section{Introduction}

With the advancement of Natural Language Processing (NLP) technology, linguistic steganography has rapidly evolved, particularly in the realm of generative steganography methods \cite{fang2017generating},\cite{li2022text},\cite{mei2022generation},\cite{yang2018rnn}.These methods are capable of autonomously generating natural text without relying on an original carrier text, exhibiting high embedding efficiency and robust anti-detection capabilities. However, the malicious exploitation of such steganography techniques can lead to serious cybersecurity issues \cite{bogdanoski2015steganography}.

In recent years, with the continuous advancement of deep learning theory and natural language processing technology, text steganalysis methods based on deep networks have emerged. These methods leverage advanced models such as Convolutional Neural Networks(CNN)\cite{wen2019convolutional}, Bi-Directional Long Short-Term Memory networks(Bi-LSTM)\cite{yang2020linguistic}, and Bidirectional Encoder Representations from Transformers(BERT)\cite{peng2021real} to automatically extract key features from the carrier. The emergence of these methods breaks the limitations of relying on prior knowledge and is more efficient in extracting semantic features compared to earlier methods.\cite{yang2022linguistic}

In the past, these deep learning-based steganalysis methods have demonstrated tremendous potential in text steganalysis. However, their success hinges on having sufficient labeled training instances for each class. The daunting task and high costs associated with reliably annotating massive datasets across various application domains often deter progress. Furthermore, the learned classifier is constrained by the coverage of its training data and can only classify instances belonging to classes encountered during training\cite{quinonero2008covariate}. However, once confronted with unseen classes, the classifier proves inadequate and lacks the capability to handle them. Such scenarios are not uncommon in practical applications. It is possible to find that training instances for each class are scarce, and it may even occur that the testing dataset contains classes not covered during training. These challenges undoubtedly pose significant hurdles to the accuracy and generalization ability of the classifier.To address this scenario and improve the performance of text steganalysis methods under zero-shot conditions, Xue et al.\cite{xue2022domain} proposed a novel text steganalysis approach based on multi-domain adaptation (MDA) in 2022, enabling the model to learn domain-invariant text features, thus enhancing detection performance on the target domain. In 2023, Xue et al.\cite{xue2023adaptive} employed an adaptive weight selection network and a feature alignment strategy based on steganographic domain distance measurement (SDDM). Fundamentally, these methods aim to mitigate the impact of domain shift by learning domain-invariant features. However, in practice, domain invariance may adversely affect discriminative capabilities.

The core idea of our approach is to adapt the model to different domains, where the weights should be correlated but different between the two domains. Rozantsev et al.\cite{rozantsev2018beyond} demonstrated in their study that partially shared weights can effectively achieve adaptability in both supervised and unsupervised settings. In this paper, our focus lies in effectively implementing steganalysis in zero-shot scenarios. We propose a method for zero-shot language steganalysis (PDTS) based on a pre-trained language model with progressive sampling pseudo-labels. Labeled data from the source domain and unlabeled data from the target domain are used for model training. In the first stage of training, the pre-trained language model is fine-tuned using labeled data from the source domain to extract initial universal language steganalysis features. In the second stage of training, self-training is conducted using unlabeled data from the target domain\cite{li2021generalized}. Pseudo-labeled samples beneficial for steganalysis are selected based on uncertainty estimation, thereby extracting more prominent and domain-adaptive language steganalysis features. Unlike the static sampling strategy employed in existing pseudo-label selection methods, we adopt a progressive sampling approach, gradually increasing the number of selected pseudo-labeled candidate samples, thereby steadily improving the model's discriminative ability through incremental learning. Finally, extensive experimental results demonstrate that the proposed method surpasses previous models in zero-shot language steganalysis, better adapting to more realistic and complex data situations, validating the reliability of our approach.

The rest of this paper is organized as follows. Section 2 introduces the related work on Cross-Domain linguistic steganalysis, attention mechanism, and pseudo-labeling. Section 3 elaborates on the proposed PDTS method. Subsequently, Section 4 presents experimental settings and evaluation results from various perspectives, followed by comprehensive discussions. Finally, Section 5 draws conclusions.

%
%
%
%


\section{Related work}
\subsection{Cross-Domain Text Steganalysis}
Domain adaptation is a particular case of transfer learning that leverages labeled data from one or more related source domains to train a text steganalysis model for unlabeled or unseen data in the target domain\cite{pan2009survey}\cite{sun2015survey}. In the field of natural language processing, the differences in data distributions, known as domain shift, are common in real-life scenarios. Hence, numerous studies focus on steganalysis of cross-domain text data. Recent work has focused on transferring deep neural network representations from labeled source datasets to target domains with sparse or absent labels. In the case of unlabeled target domains (the focus of this paper), the main strategy is to guide feature learning by minimizing the difference between the feature distributions of the source and target domains.

In 2022, Xue et al.\cite{xue2022domain} proposed a novel text steganalysis method that utilizes multi-domain adaptation (MDA) techniques to address domain mismatch issues and enhance the detection performance of steganographic texts. This method employs a pre-trained BERT model and a CNN module composed of three layers with different kernel sizes to extract domain-invariant features for each source-target domain pair and train classifiers for each domain pair. The model effectively leverages data from multiple source domains and directly optimizes the network to adapt to the target domain. However, despite the availability of various data sources in the real world, increasing the collection of labeled samples from multiple source domains does not necessarily imply more favorable transferability. This also introduces new challenges, which may result in performance inferior to that of using a single source domain.

In 2023, Xue et al.\cite{xue2023adaptive} introduced two key techniques to address challenges in cross-domain analysis. Firstly, they employed an adaptive weight selection network, which dynamically assigns weights to different features based on the distribution discrepancy between the source and target domains, effectively aligning high-dimensional feature representations. Secondly, they utilized a feature alignment strategy based on steganographic domain distance metrics, comprising the Maximum Mean Discrepancy (MMD)\cite{gretton2012kernel} and Correlation Alignment (CORAL)\cite{sun2016deep}, capable of capturing the subtle effects of weak noise generated during the information embedding process, thereby better aligning the features of the source and target domains. The combination of these two techniques forms the Cross-Domain Linguistic Steganalysis Model (SANet) proposed in the paper.Although learning domain-invariant features helps reduce the impact of domain shift, overly emphasizing this aspect may cause the model to overlook certain task-relevant domain-specific information, leading to a decrease in model performance.

\subsection{Attention Mechanism in Text Steganalysis}
The attention mechanism, originally introduced in natural language processing to address certain issues with recurrent neural network structures, has now been successfully applied to various natural language processing tasks\cite{bahdanau2014neural}. These tasks include machine translation\cite{bahdanau2014neural} , question answering\cite{sukhbaatar2015end} , document classification\cite{yang2016hierarchical} , and sentiment analysis\cite{tang2016aspect} . The intuition behind the attention mechanism is that each low-level position contributes differently to the importance of the high-level representation.

In text steganalysis, attention mechanisms are widely employed. Wen et al.\cite{wen2022few} apply attention mechanisms to task-specific feature extractors to capture subtle probability variations between different tasks. Xue et al.\cite{xue2023extraction} connect attention mechanisms with general feature extractors to enhance domain-invariant extraction of weak steganographic features, focusing on beneficial parts for steganographic feature extraction. Yang et al.\cite{yang2021sesy} utilize multi-head attention mechanisms to capture long-distance correlations and multi-aspect text understanding. Because attention mechanisms enable models to focus more on key parts of input data, they enhance the model's understanding and performance in tasks. By dynamically allocating weights to different input information, attention mechanisms allow models to focus more on important information and ignore unimportant parts, thereby improving model performance. Models incorporating attention mechanisms achieve state-of-the-art results in the aforementioned tasks and many others.

\subsection{Self-training and pseudo-label}
Pseudo-labeling is a form of self-training commonly used in semi-supervised learning with limited labeled data or in unsupervised learning with unlabeled data.\cite{triguero2015self} In this technique, pseudo-labels are initially assigned to unlabeled data based on predictions made by models trained on clean datasets. Subsequently, the algorithm iteratively reuses these data to retrain the model and update the pseudo-labels. Typically, during self-training, the predicted confidence of pseudo-labeled samples is compared to a predefined threshold. Samples with higher confidence exceeding the fixed threshold are then selected for subsequent training. Throughout the iterations, these algorithms consistently select a fixed and large number of pseudo-labeled data. However, maintaining a fixed threshold throughout the process is inappropriate. In such cases, due to the extremely limited number of training samples, the initial model may not be robust enough.\cite{chen2011co} In the initial stages, only a few pseudo-label predictions are reliable and accurate. Selecting the same quantity of data, including many unreliable predictions, would inevitably hinder the model's subsequent improvement. Hence, it's advisable to start with a small subset of pseudo-label traces, including only the most reliable and straightforward ones.\cite{bengio2009curriculum} In subsequent stages, it gradually selects more pseudo-label trace subsets to incorporate more challenging and diverse data.

\section{The proposed method}

\begin{figure}[htbp]
    \centering
    \includegraphics[width=\textwidth]{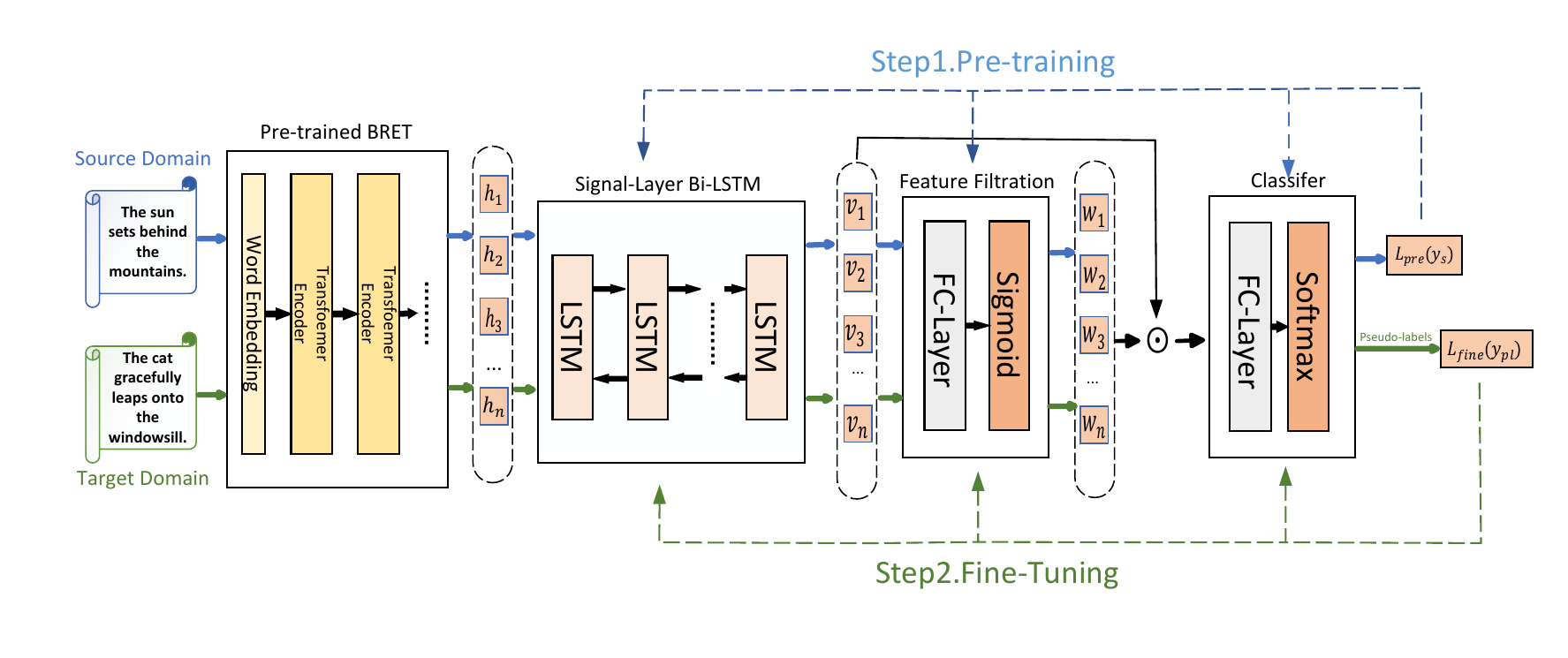}
    \caption{\scriptsize The overall structure of PDTS for cross-domain text steganalysis consists of two main stages. The first stage, represented by the blue line, involves pre-training using labeled source domain data to initialize model parameters. Following this, the second stage, indicated by the green line, is the fine-tuning process. Here, the pre-trained model generates pseudo-labels for unlabeled target domain data. These refined pseudo-labeled data are then used to further fine-tune the model, enhancing its adaptability and accuracy on target domain data.}
    \label{fig:yourlabel1}
\end{figure}

In this section, we will introduce the proposed PDTS text steganalysis model. The overall framework of the proposed network (PDTS) is depicted in Figure \ref{fig:yourlabel1}. PDTS consists of three components: a domain-agnostic feature extractor, a domain-specific feature extractor, and a classifier. Initially, domain-agnostic features are extracted from the source and target domains using pre-trained BERT. Subsequently, the domain-specific feature extractor and classifier are trained using labeled data from the source domain to establish correlation between the source and target domains. Then, the model is utilized to predict values for unlabeled data in the target domain, followed by the adoption of a progressive sampling strategy to select pseudo-labels and retrain the model, enabling it to learn useful features from the target domain.

\subsection{Preliminary}

In zero-shot learning, there are labeled training instances in the feature space, corresponding to categories known as "seen classes." Meanwhile, there are also unlabeled test instances in the feature space, belonging to another set of categories known as "unseen classes." The following defines some symbols needed: Let $S=\left \{y_{i}^{s} |i=1,...,N_{s}  \right \}$ denote the set containing a total of $ N_{s} $ seen classes, where each $y_{i}^{s}$ is a seen class.Let $T=\left \{y_{i}^{t} |i=1,...,N_{t}  \right \}$ denote the set containing a total of $ N_{t} $ unseen classes, where each $y_{i}^{t}$ is an unseen class. Note that there is no intersection between $S$ and $T$ , i.e., $S\cap T = \oslash $.Let $\chi $ represent the feature space,typically a d-dimensional real space, usually denoted as ${R^{d} }$.Let $D^{s}=\left \{ \left ( x_{i}^{s}, y_{i}^{s}\in \chi \times S \right )  \right \},i=1,...,N_{sr}    $ be the training dataset, containing $ N_{sr} $ labeled training instances,each instance $\left ( x_{i}^{s}, y_{i}^{s}\right)$  is in the feature space $\chi $,where $x_{i}^{s}$ is an instance in the feature space and $y_{i}^{s}$ is the corresponding class label.Let $X^{T}=\left \{  x_{i}^{t}\in \chi  \right \},i=1,...,N_{ta}    $ be the test dataset,containing $N_{ta}$ test instances, each $x_{i}^{t}$ is a test instance in the feature space $\chi$. And let $Y^{t}=\left \{  y_{i}^{t}\in T  \right \},i=1,...,N_{ta}  $ be the label set corresponding to the test dataset, where each $y_{i}^{t}$ is an unseen class label in the set, and these labels are the ones the model needs to predict.\cite{wang2019survey}

\begin{definition}

Zero-Shot Learning:Zero-Shot Learning is a machine learning paradigm where, from a set of labeled training instances $D_{i}^{s}$ belonging to known classes $S$,the objective is to train a classifier $f^{u}\left ( \cdot \right ) :X\to T$ that can accurately classify testing instances 
$X^{t}$ from previously unseen classes $T$ , predicting their corresponding labels $Y^{t}$ .

\end{definition}

\subsection{Domain-common Feature Extractor}
We adopted a pre-trained BERT\cite{devlin2018bert} model consisting of 12 Transformer encoder layers to extract features from both the source and target domains. Each layer is equipped with a hidden size of 768, ensuring efficiency and accuracy in feature extraction.

Zero-shot learning aims to train a classifier using known class samples to accurately classify unknown class test samples, without overlap between training and testing label spaces. This necessitates the model to generalize knowledge from known class samples to unknown ones. BERT (Bidirectional Encoder Representations from Transformers) efficiently extracts versatile features applicable to various natural language processing tasks, bolstering model performance across text processing tasks. Its effectiveness in feature extraction stems from its bidirectionality, large-scale pretraining on text data, and contextual embeddings. By considering contextual information surrounding words, BERT comprehensively understands word meanings and contexts, yielding more universal feature representations. Pretraining on extensive text data equips BERT with rich language knowledge and semantic information, enhancing its generalization capability. Furthermore, BERT utilizes contextual word embeddings, incorporating surrounding context to improve feature representation accuracy and expressive power.

Given an input sequence $X=\left \{ x_1,...,x_{len}  \right \}$, BERT transforms this input sequence into a contextual representation through a series of processing layers, such as embedding layers and a stack of $L$ transformer layers. This contextual representation encapsulates information from the entire sequence and has a fixed dimension. During this process, the embedding layer, consisting of word embeddings, positional embeddings, and token type embeddings, is responsible for converting each word into a vector representation, while the transformer layers contribute to capturing the relationships between words.
\begin{equation}
H^{0}=Embedding\left ( X \right )  
\end{equation}
\begin{equation}
H^{l}=Transformer \left ( H^{l-1}  \right ) ,l\in \left \{ 1,...,len \right \}  
\end{equation}

the output sentence representation can be written as 
$ H^{L}=\left \{ h_{1}^{L},...,h_{len}^{L}  \right \} \in R^{len\times d_{h} }$
,which will be fed into the task-specific extractor, $d_{h}$ represents the dimension of the feature vector.Please note that during the training phase, the parameters of BERT remain fixed.

\subsection{Domain-specific feature extractor}

To precisely capture the specific features of the target domain, we designed a feature filtering network composed of a single-layer bidirectional long short-term memory (Bi-LSTM) network\cite{schuster1997bidirectional} and a fully connected layer activated by the sigmoid function.

Zero-shot learning is a form of transfer learning, distinguished from other forms primarily by the heterogeneity of label spaces. To address zero-shot learning problems, it is typically necessary to utilize auxiliary information containing semantic knowledge about unknown classes, associating it with samples in the feature space to assist the learning model in better understanding the features of unknown classes and achieving accurate classification. Therefore, it is imperative to extract task-specific features from each task. These features can help the model better understand and address the specific requirements and contexts of each task. By extracting specific features from each task, the model can more effectively learn the differences between tasks and improve its discriminative power.

Bi-LSTM can simultaneously integrate past and future information into the output. Due to its effectiveness in capturing bidirectional dependencies in sequential data, it has been widely applied in various tasks such as time series prediction and natural language processing. However, in deep learning models, as features transition from general to specific and domain differences increase, the transferability of features significantly decreases at higher levels. Therefore, to address this issue, we opt to utilize a single-layer Bi-LSTM to extract domain-specific features.

Assuming we use the word representations  $h_{t}^{L} $ obtained from BERT as the input for the task-specific feature extractor. At time step $t$ ,the update of the hidden state of the Bi-LSTM is as follows:
\begin{equation}
\overrightarrow{V_{t} } =\overrightarrow{LSTM} \left (H_{t} ,H_{t-1}   \right ) 
\end{equation}
\begin{equation}
\overleftarrow{V_{t}} = \overleftarrow{LSTM} \left (H_{t} ,H_{t+1}   \right ) 
\end{equation}
\begin{equation}
V_{t} =\overrightarrow{V_{t} } \oplus \overleftarrow{V_{t} } 
\end{equation}

In fact, not all feature values in the embedding vector contribute positively to a specific task. In the context of transfer learning, in order to better differentiate between source and target samples, the discriminator may mistakenly include weak transfer features or non-transferable features (such as language style features of text) in its criteria. This may weaken the discriminator's ability to analyze steganographic text. Therefore, we introduce a feature filtration network to weight the feature embeddings, enhancing the importance of task-relevant features while reducing the influence of irrelevant or distracting features.

The feature filtration network takes ${V_{t} }$ as input and generates attention weights $w$. Its structure is very simple, consisting of a fully connected layer activated by the sigmoid function.
For text data, the weighted feature embedding $g$ is obtained through a point-wise multiplication operation between the attention weights $w$ and the input embedding ${V_{t} }$, expressed as:
\begin{equation}
w=sigmoid\left ( W_{fc}\cdot V_{t} +  b_{fc}    \right )  
\end{equation}
\begin{equation}
g = w\odot V_{t}   
\end{equation}

$W_{fc}$ and $b_{fc}$ represent the weights and biases of the fully connected layer.This design is simple yet effective, enhancing the model's generalization capability. With a straightforward fully connected layer and sigmoid activation function, the feature filtration network can adaptively adjust the importance of input features, thereby better capturing critical features of the data and improving the model's performance across various tasks and datasets.

\subsection{Classifier}
During the training phase, we pass the weighted feature vector $g$ to the linear classifier $C\left ( \cdot  \right ) $ . The output values are normalized by $softmax()$ to obtain the probability distribution $ pred$ , which can be represented as:
\begin{equation}
pred = softmax(C(g))
\end{equation}

The model's predicted output reflects the estimation of the text distribution, which represents the model's prediction of the category to which the input text belongs. By minimizing the cross-entropy between the classifier output and the true/pseudo labels, the network parameters are updated to obtain an effective classifier. This approach ensures that the model's output closely approximates the true labels, thereby enhancing the model's classification accuracy. Finally, when using batch training, the loss function is defined as follows:
\begin{equation}
Loss_{cls}=-\frac{1}{N} \sum_{i}^{} \left [ y_{i} \cdot log\left ( p_{i}  \right )+  (1-y_{i}) \cdot log\left ( 1-p_{i} \right )   \right ] 
\end{equation}
In this context, $N$ represents the batch size, $y_{i}$ denotes the true or pseudo label of sample $i$ ,and $p_{i}$ represents the predicted probability.

\begin{figure}[htbp]
    \centering
    \includegraphics[width=\textwidth]{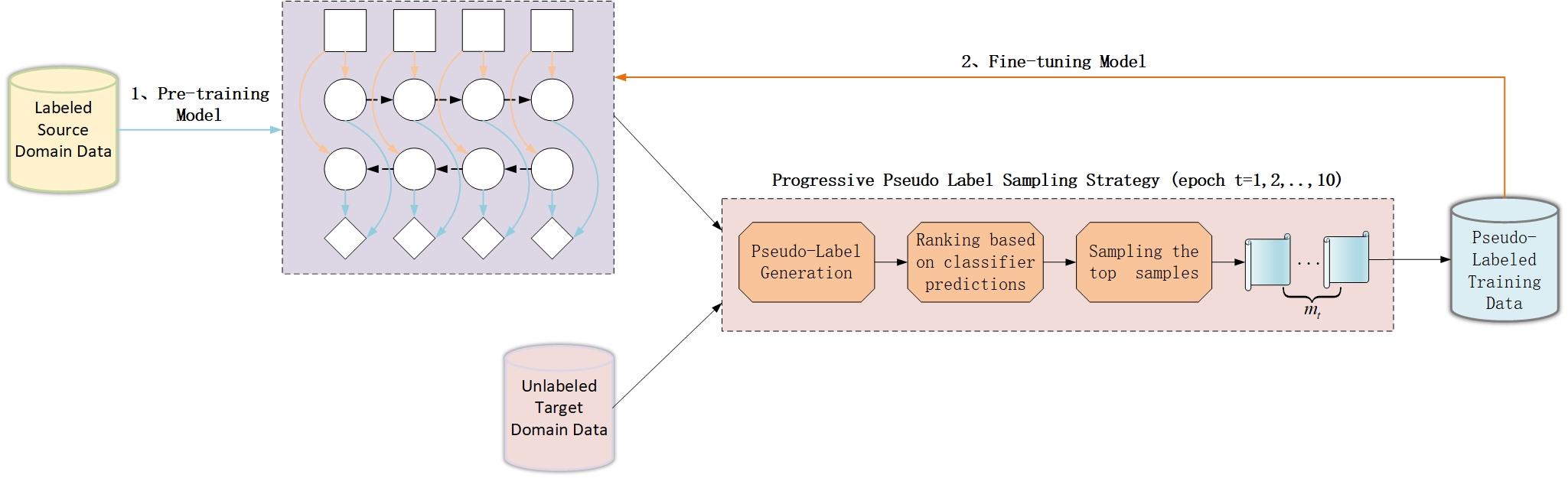}
    \caption{\scriptsize Illustration of the two-stage process of pre-training and fine-tuning.The 'epoch' in the figure refers to the number of times pseudo-labels are selected and used to fine-tune the model. $m_t$ refers to the number of pseudo-labels selected in each round of fine-tuning training, which gradually increases with each training round.}
    \label{fig:yourlabel2}
\end{figure}

\subsection{Model training}

The training process is divided into two stages: pre-training using labeled data from the source domain, and fine-tuning using unlabeled data from the target domain,as shown in Figure 1.The general outline of the training process is illustrated in Figure \ref{fig:yourlabel2}.

In the pre-training stage, labeled data $D^{s}=\left \{ \left ( x_{i}^{s}, y_{i}^{s}\in \chi \times S \right )  \right \},i=1,...,N_{sr} $ from the source domain is used as the training dataset. Each instance $\left ( x_{i}^{s}, y_{i}^{s}\right)$ in the feature space consists not only of features $ x_{i}^{s}$ but also of the corresponding class label $y_{i}^{s}$. The model parameters are optimized using cross-entropy as the loss function.

\begin{equation}
Loss_{pre-train}=-\frac{1}{N_{b} } \sum_{i}^{} \left [ y_{i}^{s}  \cdot log\left ( p_{i}^{s}  \right )+  (1-y_{i}^{s})log\left ( 1-p_{i}^{s} \right )   \right ] 
\end{equation}

Where $p_{i}^{s}$ denotes the predicted probability of the $i$ sample in the source domain, with $N_{b}$ representing the batch size during training.

After the pre-training phase, the model is fine-tuned using unlabeled data $X^{T}$ from the target domain to learn domain-specific features and improve the discriminative power of the classifier. Although the target domain data lacks labels, the model first classifies all unlabeled data. Specifically, for each unlabeled sample, we choose the class with the highest predicted probability as its pseudo-label. Then, we sort the entire dataset based on the highest predicted probabilities in descending order and select the top $m_{t} $ samples as the training data $D^{T}$ for fine-tuning. Assigning their pseudo-labels $\hat{y} $ as the true labels, denoted as $D^{T}=\left \{ \left ( x_{i}^{t}, \hat{y}\in \chi \times S \right )  \right \},i=1,...,N_{ta} $. During the fine-tuning process, the model parameters are optimized using cross-entropy as the loss function.

\begin{equation}
Loss_{fine-tune}=-\frac{1}{N_{b} } \sum_{i}^{} \left [ \hat{y}  \cdot log\left ( p_{i}^{t}  \right )+  (1-\hat{y})log\left ( 1-p_{i}^{t} \right )   \right ] 
\end{equation}

Where $p_{i}^{t}$ denotes the predicted probability of the $i$ sample in the target domain, with $N_{b}$ representing the batch size during training.

Furthermore, by setting $m_{t} = m_{t-1} +  p\cdot v$, the size of the pseudo-labeled dataset for sampling is expanded as the iteration steps $t$ increase. Here, $m_{t}$ represents the number of pseudo-labels in the t-th round,  $p\in (0,1)$ serves as an expansion factor, indicating the rate at which the training set is expanded during the iteration process.$p$ is a hyperparameter that can be set according to the task and dataset.  By increasing $m_{t}$ , candidate data for sampling is added at each iteration step to aid the model in better learning. Additionally, at each iteration step, the model is evaluated on the validation set, and the best-performing model is outputted. This helps monitor the model's performance during training and select the best-performing model for subsequent use or evaluation.

When choosing the expansion factor $p$ ,it should not be too large. If $p$ is set to a very large value, this will force the number of pseudo-labeled samples $m_t$ to increase rapidly with each iteration. However, this aggressive strategy may result in too much noise in the sampled data, leading to insufficient quality of the pseudo-labels, making them unreliable and unable to train a robust model. On the other hand, $p$ should not be too small. If $p$ is set to a very small value, the number of pseudo-labeled samples $m_t$ will increase gradually with a small step size in each iteration. The selected samples will be simple and contain relatively little information, providing almost no improvement to the model's performance. Therefore, $p$ should neither be too large nor too small. We set $p$ to 0.1 because this value strikes a balance between aggressive and conservative strategies. This setting allows the number of pseudo-labeled samples $m_t$.to increase at a moderate pace with each iteration, avoiding the high noise and low-quality pseudo-labels that result from rapid increases, as well as the insufficient information from very slow increases. This balance helps the model steadily improve performance during training while reaching satisfactory performance within a reasonable number of stages.

\section{Experiments}
\subsection{Experimental Setup}

In the experiments, we evaluated the proposed PDTS model on three widely used datasets: Twitter (T)\cite{go2009twitter}, Movie (M)\cite{maas2011learning}, and News (N)\cite{Kaggle2017news}.These datasets cover text content from various domains, including tweets, comments from social media, movie-related reviews, plot summaries, and news articles. Each dataset has its own data distribution, and all three datasets consist of common text data found in daily life. Therefore, conducting cross-domain text steganalysis across these three domains is practically meaningful. In each domain, we have two types of text: cover text and stego text. The cover text originates from three corpora, while the stego text is generated using the Conditional Probability-based Dynamic Encoding LSTM proposed by Yang et al., referred to as VLC\cite{yang2018rnn}.

To comprehensively evaluate the effectiveness of the proposed PDTS, we conducted cross-corpus experiments at five different embedding rates (bpw = 1,2,3,4,5). The training dataset for each domain comprises two parts: 10,000 cover texts (original text) randomly selected from the specified corpus and 10,000 stego texts (hidden text) generated using the same corpus. Both the validation and test datasets consist of 1,000 cover texts and 1,000 generated stego texts. It is important to note that there should be no overlap between the cover texts (original text) in the training, validation, and test datasets.

In the proposed model, we utilize the open-source BERT-based-uncased\cite{devlin2018bert} model as the shared feature extractor. To preserve its powerful feature extraction capabilities, we do not optimize the parameters of the BERT layer during training. The number of hidden units in the single-layer BiLSTM is set to 500, and the dimensionality of the word embeddings is 768. The learning algorithm employs mini-batch gradient descent with the Adam optimization algorithm, and the initial learning rate is set to 5e-5. The retention probability of the dropout layer is 0.5, and the batch size is 16. In the pre-training phase, we conducted 50 epochs, and in the fine-tuning phase, the training was performed for 10 epochs.We set $p$ to 0.1.

\subsection{Compare with the cross-domain text steganography method}

In this section, we compare the proposed PDTS with existing cross-domain text steganalysis methods, including MDA\cite{xue2022domain} and SANet\cite{xue2023adaptive}, to evaluate PDTS's effectiveness. In cross-corpus scenarios, we designed six different cross-domain steganalysis tasks (source domain $\Rightarrow$ target domain): $M\Rightarrow N$, $M\Rightarrow T$, $N\Rightarrow M$,$N\Rightarrow T$, $T\Rightarrow M$, and $T\Rightarrow N$. The experimental results, as shown in Table \ref{tab:1}, demonstrate that PDTS outperforms other methods in the testing phase.

Upon analyzing the experimental data, we found significant advantages of the PDTS model across different embedding rates, with higher accuracy and F1 scores, especially at higher embedding rates. In contrast, the SANet model showed some stability but exhibited fluctuations at certain embedding rates. Conversely, the MDA model performed relatively poorly across all embedding rates, with generally lower accuracy and F1 scores.

\begin{table*}[t]

\centering
\caption{Cross-Corpus Experiment Results Evaluating the Effectiveness of Various Steganalysis Methods against VLC Steganographic Algorithm at Multiple Embedding Rates}
\label{tab:1}
\resizebox{\textwidth}{!}{
\normalsize
\renewcommand\arraystretch{1.5}
\begin{tabular}{|c|c|c|c|c|c|c|c|c|c|c|c|c|c|c|c|c|} 
\hline
\multicolumn{2}{|c|}{Cross-Corpus}& 
\multicolumn{2}{|c|}{$M\Rightarrow N$}&
\multicolumn{2}{|c|}{$M\Rightarrow T$}&
\multicolumn{2}{|c|}{$N\Rightarrow M$}&
\multicolumn{2}{|c|}{$N\Rightarrow T$}&
\multicolumn{2}{|c|}{$T\Rightarrow M$}&
\multicolumn{2}{|c|}{$T\Rightarrow N$}&
\multicolumn{2}{|c|}{Average}\\
\hline

\makebox[1.5cm]{Model} & bpw & ACC & F1& ACC & F1& ACC & F1& ACC & F1& ACC & F1& ACC & F1& ACC & F1 \\
\hline

\multirow{5}{*}{MDA} & 1 &0.8646 &0.8630 &0.7396 &0.7345 &0.7396 &0.7231 &0.7188 &0.7125 &0.7292 &0.7184 &0.8438 &0.8405 &0.7726 &0.7654 \\
\cline{2-16}

\multirow{5}{*}{}& 2 &0.8542 &0.8516 &0.7292 &0.7255&0.6875&0.6631&0.7604&0.7527&0.7292&0.7135&0.8125&0.8104&0.7621 &0.7528 \\
\cline{2-16}
\multirow{5}{*}{}& 3 &0.8229&0.8187&0.6875&0.6762&0.7457&0.7345&0.7083&0.6949&0.6832&0.6645&0.7604&0.7585&0.7346 &0.7246 \\
\cline{2-16}
\multirow{5}{*}{}& 4 &0.7812 &0.7816&0.7028&0.6956&0.6544&0.6275&0.6458&0.6361&0.6624&0.6428&0.8021&0.7992&0.7081 &0.6971 \\
\cline{2-16}
\multirow{5}{*}{}& 5 &0.7396&0.7401&0.6560&0.6366&0.6562&0.6198&0.6564&0.6316&0.6875&0.6605&0.7812&0.7768&0.6961 &0.6775 \\
\hline

\multirow{5}{*}{SANet} &1&0.8802&0.8799&0.8002&0.7997&0.8742&0.8732&0.8163&0.8159&0.8595&0.8583&0.8628&0.8625&0.8488 &0.8482\\
\cline{2-16}
\multirow{5}{*}{}& 2 &\textbf{0.8577}&\textbf{0.8576}&0.7903&0.7902&0.8412&0.8394&0.7890 &0.7887&0.8498
&0.8485&0.8575&0.8572&0.8309 &0.8302 \\
\cline{2-16}
\multirow{5}{*}{} & 3 &0.8205&0.8202&0.7757&0.7756&0.8223&0.8197&0.7702&0.7694&0.8108&0.8075
&0.8203&0.8192&0.8033 &0.8019 \\
\cline{2-16}
\multirow{5}{*}{} & 4 &0.7945&0.7940 &0.7478&0.7472&0.7878&0.7852&0.7502&0.7490 &0.7570 &0.7477&0.7745&0.7706&0.7686 &0.7655 \\
\cline{2-16}
\multirow{5}{*}{} & 5 &0.7650 &0.7649&0.7263&0.7256&0.7505&0.7452&0.7218&0.7204&0.7373&0.7254&0.7487&0.7435&0.7416 &0.7375 \\
\hline

\multirow{5}{*}{PDTS} & 1 & \textbf{0.8945} & \textbf{0.8940} & \textbf{0.8115} & \textbf{0.8111} & \textbf{0.8930} & \textbf{0.8912} & \textbf{0.8215} & \textbf{0.8206} & \textbf{0.8995} & \textbf{0.8987} & \textbf{0.8685} & \textbf{0.8682} & \textbf{0.8648} & \textbf{0.8640} \\
\cline{2-16}
\multirow{5}{*}{} & 2 & 0.8535 & 0.8522 & \textbf{0.8045} & \textbf{0.8029} & \textbf{0.8622} & \textbf{0.8601} & \textbf{0.8121} & \textbf{0.8118} & \textbf{0.8655} & \textbf{0.8648} & \textbf{0.8760} & \textbf{0.8747} & \textbf{0.8455} & \textbf{0.8444} \\
\cline{2-16}
\multirow{5}{*}{} & 3 & \textbf{0.8225} & \textbf{0.8210} & \textbf{0.8015} & \textbf{0.7965} & \textbf{0.8305} & \textbf{0.8291} & \textbf{0.7950} & \textbf{0.7944} & \textbf{0.8434} & \textbf{0.8422} & \textbf{0.8450} & \textbf{0.8439} & \textbf{0.8229} & \textbf{0.8212} \\
\cline{2-16}
\multirow{5}{*}{} & 4 & \textbf{0.8005} & \textbf{0.8001} & \textbf{0.7670} & \textbf{0.7654} & \textbf{0.8122} & \textbf{0.8177} & \textbf{0.7683} & \textbf{0.7672} & \textbf{0.8210} & \textbf{0.8207} & \textbf{0.8073} & \textbf{0.8059} & \textbf{0.7960} & \textbf{0.7952} \\
\cline{2-16}
\multirow{5}{*}{} & 5 & \textbf{0.7675} & \textbf{0.7671} & \textbf{0.7452} & \textbf{0.7443} & \textbf{0.7710} & \textbf{0.7672} & \textbf{0.7691} & \textbf{0.7677} & \textbf{0.7701} & \textbf{0.7768} & \textbf{0.7620} & \textbf{0.7609} & \textbf{0.7642} & \textbf{0.7640} \\
\hline

\end{tabular}
}
\end{table*}

Further observation revealed a decreasing trend in performance for all models as the embedding rate increased, especially at higher rates. However, the PDTS model maintained relatively stable performance under these conditions, with a milder decline compared to other models, demonstrating higher robustness. Across all embedding rates, the PDTS model demonstrated significantly higher average accuracy and F1 scores compared to the MDA and SANet models. This indicates that the PDTS model performs better in cross-domain text steganalysis. Particularly noteworthy is its higher robustness compared to other models when dealing with high embedding rates, further emphasizing its superiority in cross-domain text steganography tasks.

In the 2-embedding-rate M-to-N transfer, SANet achieves an ACC that is 0.42\% higher than our proposed model PDTS. This may be attributed to SANet's utilization of an adaptive weight selection network. Given the similarity between the source domain (Movie) and target domain (News), SANet shares the same weights for both domains, enhancing the influence of source domain data on the target domain classifier.

Considering the comprehensive experimental results, we conclude that the PDTS model may be the optimal choice for cross-domain text steganalysis, especially when handling high embedding rates. Its robust performance makes it highly applicable across various cross-domain text steganography tasks.

\subsection{Ablation Study}

We conducted ablation studies to quantify the impact of key components in the proposed model. At an embedding rate of 1, utilizing VLC for generating stego-text, we designed six different cross-domain steganalysis tasks. Here, "w-PL" denotes PDTS without fine-tuning using pseudo-labels, "w-FF" denotes PDTS without the feature filtering layer, and "w-SLB" denotes PDTS with multiple layers of Bi-LSTM instead of a single layer. The results of the ablation study are shown in Table \ref{tab:2}.

\begin{table*}[t]

\centering
\caption{Cross-Corpora Scenario Ablation Experiment Results}
\label{tab:2}
\resizebox{\textwidth}{!}{
\normalsize
\renewcommand\arraystretch{1.5}
\begin{tabular}{|c|c|c|c|c|c|c|c|c|c|c|c|c|c|c|c|c|} 
\hline
\multicolumn{2}{|c|}{Cross-Corpus}& 
\multicolumn{2}{|c|}{$M\Rightarrow N$}&
\multicolumn{2}{|c|}{$M\Rightarrow T$}&
\multicolumn{2}{|c|}{$N\Rightarrow M$}&
\multicolumn{2}{|c|}{$N\Rightarrow T$}&
\multicolumn{2}{|c|}{$T\Rightarrow M$}&
\multicolumn{2}{|c|}{$T\Rightarrow N$}&
\multicolumn{2}{|c|}{Average}\\
\hline

\multicolumn{2}{|c|}{Model} & ACC & F1& ACC & F1& ACC & F1& ACC & F1& ACC & F1& ACC & F1& ACC & F1 \\
\hline

\multicolumn{2}{|c|}{w-PL} &0.8355 &0.8312 &0.8082 &0.8063 &0.8755 &0.8732 &0.8183 &0.8175 &0.8734 & 0.8722&0.8525 &0.8512 &0.8439 &0.8419 \\
\hline
\multicolumn{2}{|c|}{w-FF} &0.8832 &0.8827 &0.8005 &0.8001 &\textbf{0.8965} &\textbf{0.8961} &0.8192 &0.8188 &0.8851 &0.8842&0.8662 &0.8656 &0.8584 &0.8579 \\
\hline
\multicolumn{2}{|c|}{w-SLB} &\textbf{0.8952} &\textbf{0.8945} &0.8075 &0.8077 &0.8785 &0.8772 &0.8215 &0.8198 &0.8855 &0.8852 &0.8685 &0.8679 &0.8594 &0.8587 \\
\hline
\multicolumn{2}{|c|}{PDTS} & 0.8945 & 0.8940 &\textbf{ 0.8115} & \textbf{0.8111} & 0.8930 & 0.8912 & \textbf{0.8215} & \textbf{0.8206} & \textbf{0.8995} & \textbf{0.8987} & \textbf{0.8685} & \textbf{0.8682} & \textbf{0.8648} & \textbf{0.8640} \\
\hline

\end{tabular}
}
\end{table*}

In the ablation experiments, we found that the w-PL model performed moderately in most tasks, with slightly lower average accuracy and F1 scores compared to other models. However, the data shows that the self-training approach using pseudo-labels can significantly improve the classifier's performance. As mentioned earlier, when the data distribution in the target domain differs from the source domain, using pseudo-labels can better adapt to the characteristics of the target domain data, resulting in better classification performance.

The w-SLB model performed best in the M→N task. This could be attributed to the use of a multi-layer neural network, which can obtain richer feature representations. Additionally, the MOVIE and NEWS datasets are relatively similar, making features transferred from the source domain more likely to be useful in the target domain. However, in other tasks, the performance of the w-SLB model was inferior to PDTS. This may be because while multi-layer Bi-LSTM can obtain more features, deeper feature representations are more focused on the source domain, leading to poorer classification performance in the target domain.

In experiments, it was observed that in the M-to-N experiment, the adopted multi-layer Bi-LSTM model outperforms our proposed model PDTS by 0.07\%. This may be because the multi-layer neural network extracts more features than a single-layer neural network, and since the domains Movie and News are quite similar, the pretrained model is more helpful for the target domain classifier. In the N-to-M experiment, the model that abandons the use of the feature filtering network outperforms our proposed model PDTS by 0.55\%. This might be because the feature filtering network sometimes filters out some useful features for classification.

Overall, PDTS and the w-SLB model performed best in most tasks, exhibiting higher accuracy and F1 scores. In the experiments, PDTS excelled in multiple tasks, particularly when dealing with target domains labeled as T, demonstrating its potential and superiority in cross-domain steganalysis.

\begin{figure}[htbp]
    \centering
    \begin{subfigure}{0.3\textwidth}
        \centering
        \includegraphics[width=\linewidth]{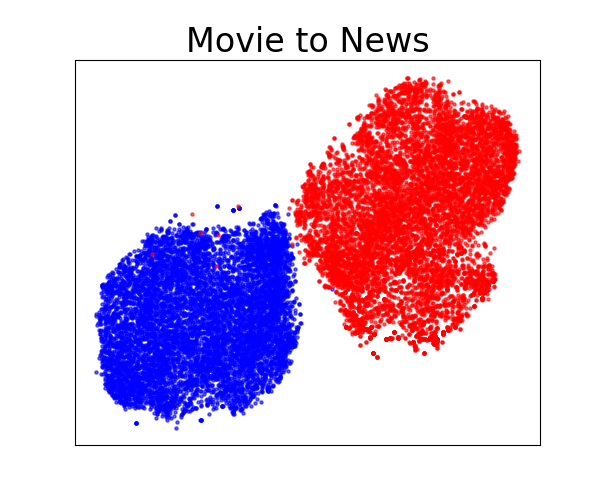}
        \label{a}
    \end{subfigure}
    \begin{subfigure}{0.3\textwidth}
        \centering
        \includegraphics[width=\linewidth]{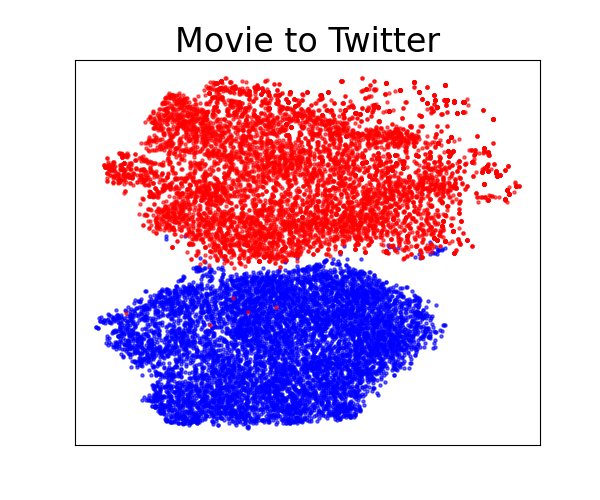}
        \label{b}
    \end{subfigure}
    \begin{subfigure}{0.3\textwidth}
        \centering
        \includegraphics[width=\linewidth]{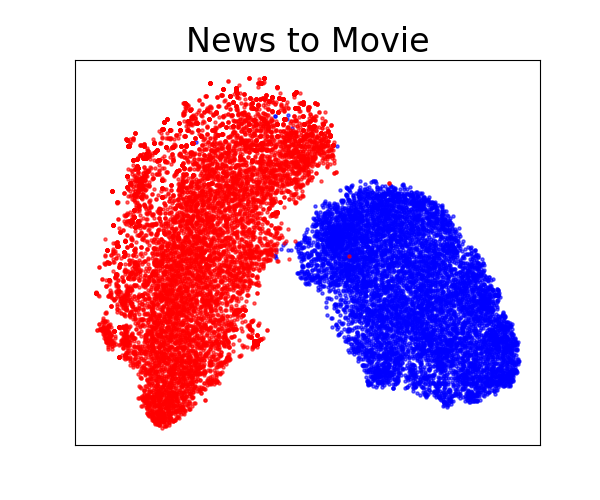}
        \label{c}
    \end{subfigure}
    \\
    \begin{subfigure}{0.3\textwidth}
        \centering
        \includegraphics[width=\linewidth]{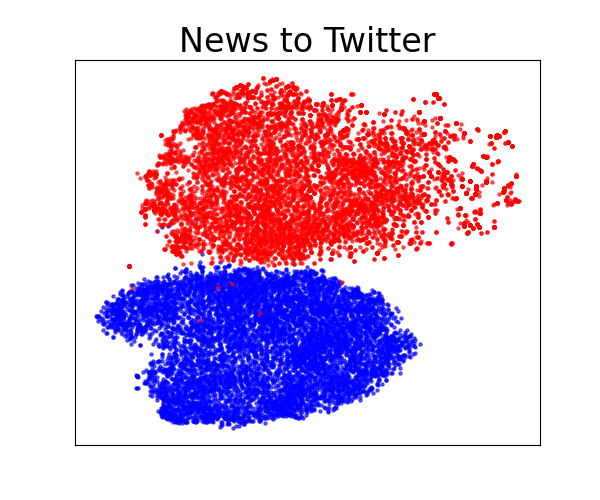}
        \label{d}
    \end{subfigure}
    \begin{subfigure}{0.3\textwidth}
        \centering
        \includegraphics[width=\linewidth]{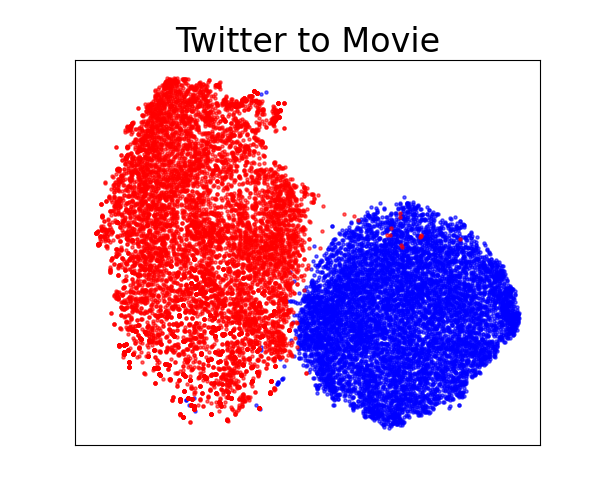}
        \label{e}
    \end{subfigure}
    \begin{subfigure}{0.3\textwidth}
        \centering
        \includegraphics[width=\linewidth]{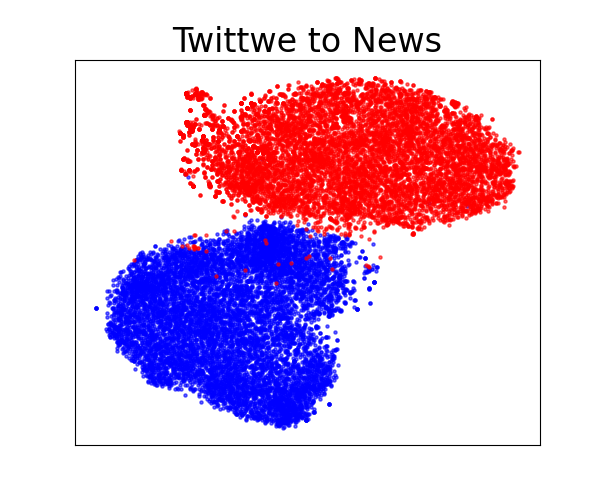}
        \label{f}
    \end{subfigure}
    \caption{
In the six cross-domain tasks, the features extracted by PDTS were visualized. In the visualization, blue dots represent cover text, while red dots represent steganographic text.}
    \label{fig:six_images}
\end{figure}

At an embedding rate of 1, we visualized the extracted features using dimensionality reduction techniques in six cross-domain tasks, as shown in Figure \ref{fig:six_images}. Each blue point in the figure represents a cover text, while each red point represents a steganographic text, with the title of each subplot indicating the respective task. From the figure, it can be observed that in all six tasks, using the fine-tuned model to extract features results in a clear boundary between cover texts and steganographic texts, which represents the decision boundary of the classifier. For each category, the feature distribution extracted by our PDTS model is highly concentrated, effectively adjusting the feature space between the source and target domains.

\section{Conclusion}

This paper addresses the cross-domain zero-shot problem by proposing a novel linguistic steganalysis framework, consisting of a domain-agnostic feature extractor and a domain-specific feature extractor. The method involves two-stage parameter updates, utilizing labeled source domain data to pre-train the model for extracting general features, and then fine-tuning the model on the target domain using pseudo-labels to acquire domain-specific features, thereby enhancing the model's classification capability in the target domain. Experimental results demonstrate that the proposed PDTS outperforms state-of-the-art cross-domain linguistic steganalysis methods. In the future, we will explore other techniques to further enhance the performance of cross-domain zero-shot steganalysis methods, such as employing adversarial neural networks.

\subsubsection{Acknowledgements} 

This work was supported by the National Natural Science Foundation of China under Grants U21B2020.

\begin{refcontext}[sorting = none]
\printbibliography
\end{refcontext}

\end{document}